\documentclass[a4paper, 10pt, conference]{ieeeconf}      
\usepackage{FG2024}

\FGfinalcopy 

\IEEEoverridecommandlockouts                              
\usepackage{amsmath} 
\usepackage{algpseudocode} 
\usepackage{algorithm}
\usepackage{algorithmicx}
\usepackage{booktabs} 
\usepackage{multirow}
\usepackage{graphicx} 
\usepackage{graphics} 
\usepackage{caption}
\usepackage{xcolor,soul} 
\newcommand{\hlc}[2][yellow]{{%
    \colorlet{foo}{#1}%
    \sethlcolor{foo}\hl{#2}}%
}
\makeatletter
    \def\footnoterule{\kern-3\p@
      \hrule \@width 0.8in \kern 2.6\p@} 
\makeatother
\usepackage{pifont}
\newcommand{\cmark}{\text{\ding{51}}}
\newcommand{\xmark}{\text{\ding{55}}}

\def\FGPaperID{****} 

\title{\LARGE \bf
Improving Personalisation in Valence and Arousal Prediction using Data Augmentation 
}

\author{\parbox{16cm}{\centering
    {\large Munachiso Nwadike$^1$ Jialin Li$^1$, Hanan Salam$^1$}\\
    {\normalsize
    $^1$ Department of Computer Science, New York University Abu Dhabi, UAE }}
    \thanks{This work was supported by the Center for Artificial Intelligence and Robotics at NYUAD}
}

\begin{document}

\ifFGfinal
\thispagestyle{empty}
\pagestyle{empty}
\else
\author{Anonymous FG2024 submission\\ Paper ID \FGPaperID \\}
\pagestyle{plain}
\fi
\maketitle

\begin{abstract}
In the field of emotion recognition and Human-Machine Interaction (HMI), personalised approaches have exhibited their efficacy in capturing individual-specific characteristics and enhancing affective prediction accuracy. However, personalisation techniques often face the challenge of limited data for target individuals. This paper presents our work on an enhanced personalisation strategy, that leverages data augmentation to develop tailored models for continuous valence and arousal prediction. Our proposed approach, Distance Weighting Augmentation (DWA), employs a weighting-based augmentation method that expands a target individual's dataset, leveraging distance metrics to identify similar samples at the segment-level. Experimental results on the  MuSe-Personalisation 2023 Challenge dataset demonstrate that our method significantly improves the performance of features sets which have low baseline performance, on the test set. This improvement in poor-performing features comes without sacrificing performance on high-performing features. In particular, our method achieves a maximum combined testing CCC of 0.78, compared to the reported baseline score of 0.76 (reproduced at 0.72). It also achieved a peak arousal and valence scores of 0.81 and 0.76, compared to reproduced baseline scores of 0.76 and  0.67 respectively.
Through this work, we make significant contributions to the advancement of personalised affective computing models, enhancing the practicality and adaptability of data-level personalisation in real world contexts.
\end{abstract}

\section{INTRODUCTION}
Automatic emotion recognition (ER) has been an active area of research in the field of Human-Machine Interaction (HMI) during the past decade. 
ER utility spans a wide range of applications in areas like healthcare \cite{hasnul2021electrocardiogram,ayata2020emotion,pujol2019emotion}, and robotics \cite{salam2023automatic,salam2015engagement}.  Research approaches in automatic ER has focused on either the categorical model \cite{senechal2012facial,hazourli2021multi} or the dimensional model \cite{soladie2012multimodal,soladie2013continuous}. The categorical model focuses on the recognition of discrete emotion classes (anger, disgust, fear, happiness, sadness, surprise, contempt). This model of ER has been widely explored in affective computing for its simplicity, due to the ease with which humans can readily interpret distinct emotional classes. The dimensional model, on the other hand, focuses on the prediction of continuous emotional attributes representing them in a continuous dimensional space such as arousal-valence space. The dimensional model is more suitable for modeling the complex emotions, which cannot easily be fit into distinct categories, but rather vary along a spectrum. Both models of emotions have been investigated and contrasted in the affective computing literature \cite{ps2017emotion,bruna2016emotion,song2016perceived}.

ER presents various complex challenges due to the diverse nature of emotional expressions across individuals and cultures.
Each individual expresses emotion slightly differently  \cite{jensen2016personality,shaqra2019recognizing}, in a range of modalities such as voice audio, video facial expression, and physiological stress signals such as heart rate (BPM) \cite{stappen2021muse,stappen2021muse2}. The question naturally arises, as to whether personalisation strategies can be used with multimodal supervised learning techniques to improve state-of-the-art performance on dimensional emotion recognition problems. 
Efforts such as the 4th Multimodal Sentiment Analysis Challenge and Workshop: Mimicked Emotions, Humour and Personalisation (MuSe 2023) \cite{christ2023muse} have sought to address this problem. Specifically, the MuSe-personalisation sub-challenge focuses on investigating customised approaches to enhance the accuracy of machine learning for customised predictions of emotional valence and arousal. A multimodal dataset, Ulm-Trier Social Stress Test (\textsc{Ulm-TSST}) \cite{stappen2021muse}, has been provided, together with seven pre-extracted feature sets. 

Early research in affective machine learning has yielded significant insights into the potential of personalisation for predicting emotional states \cite{kim2009towards,vogt-andre-2006-improving}. Techniques such as transfer learning \cite{kathan2022personalised,Rescigno202035811,woodward2021towards,barros2022ciao,li2020personality,ren2022predicting,rudovic2018personalized}, user-specific training \cite{yusuf2017individuality,shah2021personalized,shao2021personality,leyzberg2014personalizing,abbasi2009towards,Shahabinejad2021toward}, group-specific training \cite{vogt-andre-2006-improving,kim2009towards,shaqra2019recognizing,kasparova2020inferring} and multitask learning \cite{lopez2018multi,jaques2016multi,taylor2017personalized,saeed2017personalized,chithrra2022personalized} have demonstrated notable improvements in estimation performance \cite{li2023survey}. The MuSe-Personalisation challenge baseline paper \cite{christ2023muse}, for instance, adopted the transfer learning approach. This approach involves training a generic model on global data and fine-tuning it using individualised data specific to each user. 
However, achieving effective personalisation in real-world settings remains a considerable challenge due to data constraints \cite{li2023survey}. The limited data availability and quality for target users poses significant challenges to the performance of the personalised models. To address these constraints, previous research has explored various techniques, including transfer learning \cite{rescigno2020personalized}, weighting-based approaches \cite{chattopadhyay2012multisource,chu2016selective}, feature augmentation \cite{team2iccv,zhao2018transferring,salam2016fully}, and generative-based models \cite{niinuma2022facial,alyuz2016semi,liang2020pose,barros2019personalized,yang2018identity,wang2018personalized}.
Among them, weighting-based methods offer a unique advantage by allowing the fine-tuning of weights as hyperparameters during augmentation, providing added flexibility for model training. However, traditional approaches in weighting-based methods have not fully tapped into their potential. Most methods overlook the opportunity to utilise a two-stage procedure involving training a generic model on global non-personalised data. Consequently, they do not fully exploit the abundant information contained within this global data.

Our work aims to contribute to and advance the field of multimodal personalised affective computing, by leveraging the insights of weighting-based methods. Building upon the foundations laid by previous studies, particularly the curriculum learning and grouping methods introduced in \cite{schneider2021personalization}, we propose a distance weighting augmentation (DWA) approach. DWA may be interpreted as existing at the intersection of weighting-based methods and transfer learning. By experimenting on the \textsc{Ulm-TSST} dataset using this approach, we seek to shed light on its potential and explore its effectiveness in enhancing affective understanding. The main contributions of this paper can be summarised as follows:
 
\begin{itemize} 
  \item We highlight the importance of personalisation across 8 different features in 4 different modalities, by showing the difference in model performance with and without personalisation.
  \item We introduce the DWA method, as a data augmentation strategy to enhance model performance in valence and/or arousal prediction, showing how DWA aids performance of features which are poorly performing.  
  \item Through experimentation on competing distance metrics, we illustrate how different feature modalities respond differently to DWA, and how valence and arousal signals may be improved independently of one another.  
\end{itemize} 

\section{RELATED WORK} 
This work addresses multi-modal emotion recognition and personalisation, and it is based on sample re-weighting and individual model fine-tuning approaches. In this section, we give a brief review of these techniques.

\subsection{Generic Multimodal Emotion Recognition} 
Emotion recognition refers to the field developing technical skills such as facial recognition, speech recognition, voice recognition, deep learning, and pattern recognition for recognizing human emotions.
Among all the emotion recognition tasks, arousal-valence estimation has gained a lot of attention due to its ability to capture a wide range of emotional states in a nuanced manner. As essential dimensions in affect modeling, they represent emotional polarity and intensity, respectively. The estimation of these natural emotions can be derived from biosignals collected via wearable sensors \cite{siirtola2023predicting}, text \cite{odaka2023block}, speech \cite{cunningham2019audio}, video \cite{savchenko2023emotieffnets} and so on.  

Emotion recognition from multi-modal signals has gained considerable scholarly attention in recent years \cite{ahmed2023systematic}. Generic multi-modal approaches combine more than two different modalities like speech \cite{povolny2016multimodal}, visual \cite{poria2016convolutional}, audio \cite{de2023leveraging}, text \cite{alswaidan2020survey}, and physiological signals \cite{song2021hidden} to recognize emotions. They extract informative features from each modality and fuse them, either through feature-level fusion or decision-level fusion after separate modeling per modality. 
The recent advancements in deep learning and multi-modal emotion recognition have significantly improved the accuracy of fine-grained valence-arousal estimation \cite{kollias2023abaw}.  These include but are not limited to the studies on audio-visual scenarios \cite{praveen2023audio,liu2023evaef}, and audio-textual scenarios \cite{de2023leveraging}. For example, the multi-modal fusion framework proposed by \cite{praveen2023audio} has enhanced valence-arousal estimation by leveraging complementary audio-visual modalities. The framework employs a joint cross-attentional architecture that calculates attention weights based on the correlation between combined audio-visual features and individual audio and video features, thus effectively extracts salient features across modalities while preserving intra-modal relationships. Another audio-textual framework explored multiple temporal models to incorporate sequential information in the visual signals with various ensemble strategies at decision level \cite{liu2023evaef}. On an audio-textual level, the method proposed in \cite{de2023leveraging} developed a distilled encoder using audio-text features, significantly improving valence prediction while addressing the size constraints of large self-supervised learning (SSL) models.
However, these generic models make general assumptions and cannot capture individual differences in emotional expression well. Consequently, personalised models that leverage on individual characteristics seem promising.

\subsection{Personalised Multi-modal Emotion Recognition}
Early works in personalising affective computing models present from two angles, data-level techniques and model-level techniques \cite{li2023survey}. Model-level techniques focus on directly incorporating personalisation in model architecture or training \cite{barros2022ciao,Rescigno202035811,jaques2016multi}, while data-level techniques operate on data before feeding it into the model \cite{schneider2021personalization,ren2022predicting, yusuf2017individuality,vogt-andre-2006-improving}. Prior works have notably advances by accounting individual differences in social behavior and emotional states through tailoring the generic affective models to individual-specific and group-specific levels. In \cite{zhao2019personalized}, for example, a multi-task learning approach was proposed for personalised emotion recognition from physiological signals. By modeling relationships between signals and personality in a hypergraph framework, emotion recognition was formulated as related tasks for multiple features, improving personalised emotion prediction accuracy. 

\subsubsection{Individual Model Transfer Learning}
Individual transfer learning is a two-step process that is widely adapted in personalised deep learning algorithms. It begins with the training of a deep learning model on a generic dataset, followed by fine-tuning the model on a dataset that is specific to an individual. This approach strikes a balance between generalisability and specialisation, ensuring that the datasets are utilised to their fullest extent. The research by \cite{schneider2021personalization} introduces the concept of ``late shaping," which aims to enhance the model's exposure to individual-specific data during the training phase using transfer learning. It fine-tunes a pre-trained model using the data of the individual of interest. Similarly, \cite{kathan2022personalised} proposes to adapt the pre-trained general model for individual-level humor recognition by manually fine-tuning the generic model on an individual humor dataset. In the context of facial expression recognition, \cite{barros2022ciao} also applies individual model transfer learning to facilitate the recognition of non-universal expressions. It builds on the feature extraction capabilities of a pre-trained convolutional encoder, and augments the existing encoder with an additional convolutional layer. This layer is specifically designed to learn and capture the unique representations inherent to the target dataset, thereby enhancing the model's performance in specialised facial expression recognition tasks.

\subsubsection{Sample Re-weighting}

Re-weighting methods primarily attach weights to training samples, or to personalised model predictions that come from similar user datasets, based on their similarity to the target user's data. By effectively utilising the available dataset, the model becomes better at capturing the unique characteristics of the target user, leading to more accurate predictions. In \cite{schneider2021personalization}, ``data grouping" was proposed to augment the individual's small MNIST dataset with similar data points from the global MNIST dataset, with its similarity identified using an autoencoder. 

Motivated by such existing investigations into personalisation strategies, we propose to investigate the potential of the personalised multi-modal emotion recognition using a deep network trained under a combined re-weighting \textit{and} transfer learning scheme.

\begin{algorithm}
\caption{Distance Weighting Augmentation}
\label{alg:algo1}
\scriptsize
\textbf{Input: }Global Data $D_{G}$,  Data of Individual $D_{I}$\\
\textbf{Output: }Augmented data of individual $I$
\begin{algorithmic}[1]
\Ensure All segment lengths == \textit{winlen}  
\Statex{}
\Statex{\textbf{Generate \textit{AugPool}} 
 $AP$}\Comment{Build Augmentation Pool from $D_{G}$}
\State{$AP := \{\}$}
\For{each individual $D_{G_{i}} \in D_{G}$}
  \For{each sample $S \in D_{G_{i}}$} \Comment{each sample is a segment}
    \State{$AP := AP \cup S $}

  \EndFor
\EndFor

\Statex{}
\Statex{\textbf{Perform Augmentation}}  
\For{each individual $D_{I_{i}} \in D_{I}$}
\State{$D_{I_{i}aug}:= D_{I_{i}}$}
  \For{each sample $D_{I_{i}W^{T}} \in D_{I_{i}}$}  
  
        \State $W^T :=  D_{I_{i}W^{T}} $
        
        \For{each sample $S \in AP$} 
        
        \If{$\text{distance\_metric} == \text{``cosine''}$}
        \State{$distance:= \text{cosine\_distance}(S, W^T) $}
        \State{$ \ \ \ \ \ \ \ \ \ \ \ \ \    = \displaystyle\sum_{j=1}^{winlen} [1-\frac{\overrightarrow{S_j} \cdot  \overrightarrow{W^T_j}}{\parallel\overrightarrow{S_j}\parallel^{2}\parallel \overrightarrow{W_{j}^{T}}\parallel^{2}}]$} 
        
        \ElsIf{$\text{distance\_metric} == \text{``centroid DP''}$}
        \State{ $\overrightarrow{W}:= \overline{W^T} \ ; \ \overrightarrow{S}$:= $\overline{S}$ } \Comment{centroid of both}
        \hspace{4cm}\Comment{\hspace{5.5cm}segments}
        \State{$distance:= -\overrightarrow{W} \cdot \overrightarrow{S}$} 
        
        \ElsIf{$\text{distance\_metric} == \text{``centroid L2''}$}
        \State{ $\overrightarrow{W}:= \overline{W^T} \ ; \ \overrightarrow{S}:= \overline{S}$ }
        \State{$distance:= \parallel\overrightarrow{W}-\overrightarrow{S}\parallel^{2}$}
        \EndIf
        
        \EndFor  
        
        \State{$ D_{aug} := n\ samples\ S \in AP$}
        \State{\hspace{3.0cm}$with\ \min distance$}
        \State{$ D_{I_{i}aug} :=  D_{I_{i}aug} \cup D_{aug}  $}
        
    \EndFor  
\EndFor  
\end{algorithmic}
\end{algorithm}
\section{PROPOSED METHODOLOGY}

In this section, we present a data-level augmentation approach, aimed at enhancing personalised models’ accuracy in predicting continuous valence and arousal under data label constraints. We provide a detailed explanation of our approach, organised into two sections. First, in section \ref{dwa}, we introduce the distance weighting framework. Then, in section \ref{distanc-metrics}, we discuss the criteria employed for the distance metrics used to calculate the weights during augmentation.  

\subsection{Distance Weighting Augmentation} \label{dwa}
\begin{figure}[h!]
    \begin{center}
    \includegraphics[width=\linewidth]{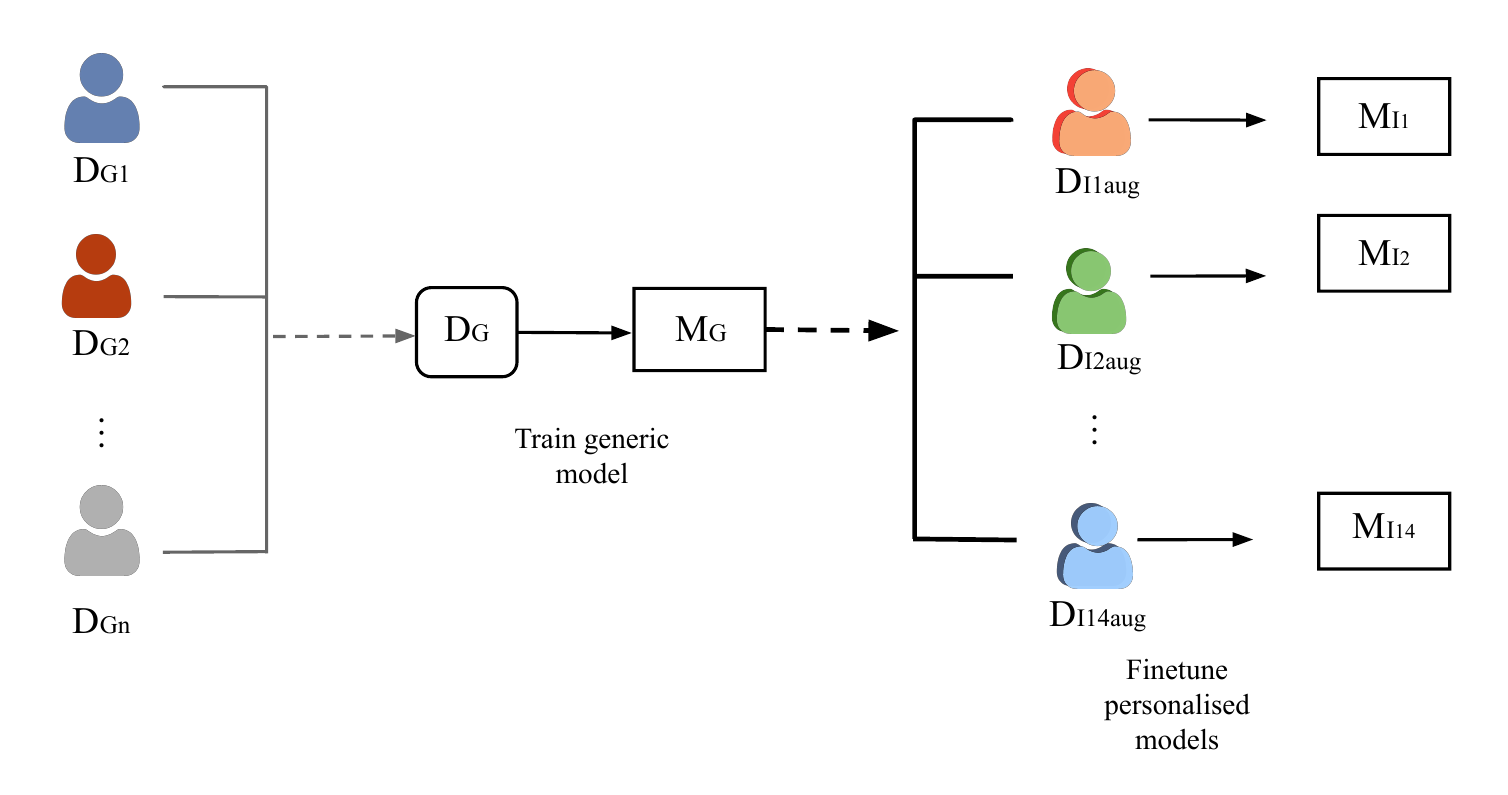}
    \end{center}
  \caption{Our personalisation framework leverages transfer learning together with distance weighting augmentation.}
\label{fig:pipeline}
\end{figure}

Our objective is to enhance emotion recognition performance, on each unique individual, with respect to our evaluation metric. Overall, we adopt a transfer learning technique as illustrated in Figure \ref{fig:pipeline}. Recognizing the importance  of the size and quality of the individual dataset $D_{I_i}$, we propose the Distance Weighting Augmentation (DWA) method. DWA seeks to augment the limited $D_{I_i}$, by incorporating similar samples from the global dataset $D_{G}$. Psuedo-code is provided in Algorithm \ref{alg:algo1} for a comprehensive breakdown of DWA.

\begin{figure}
    \begin{center}
    \includegraphics[width=0.48\textwidth]{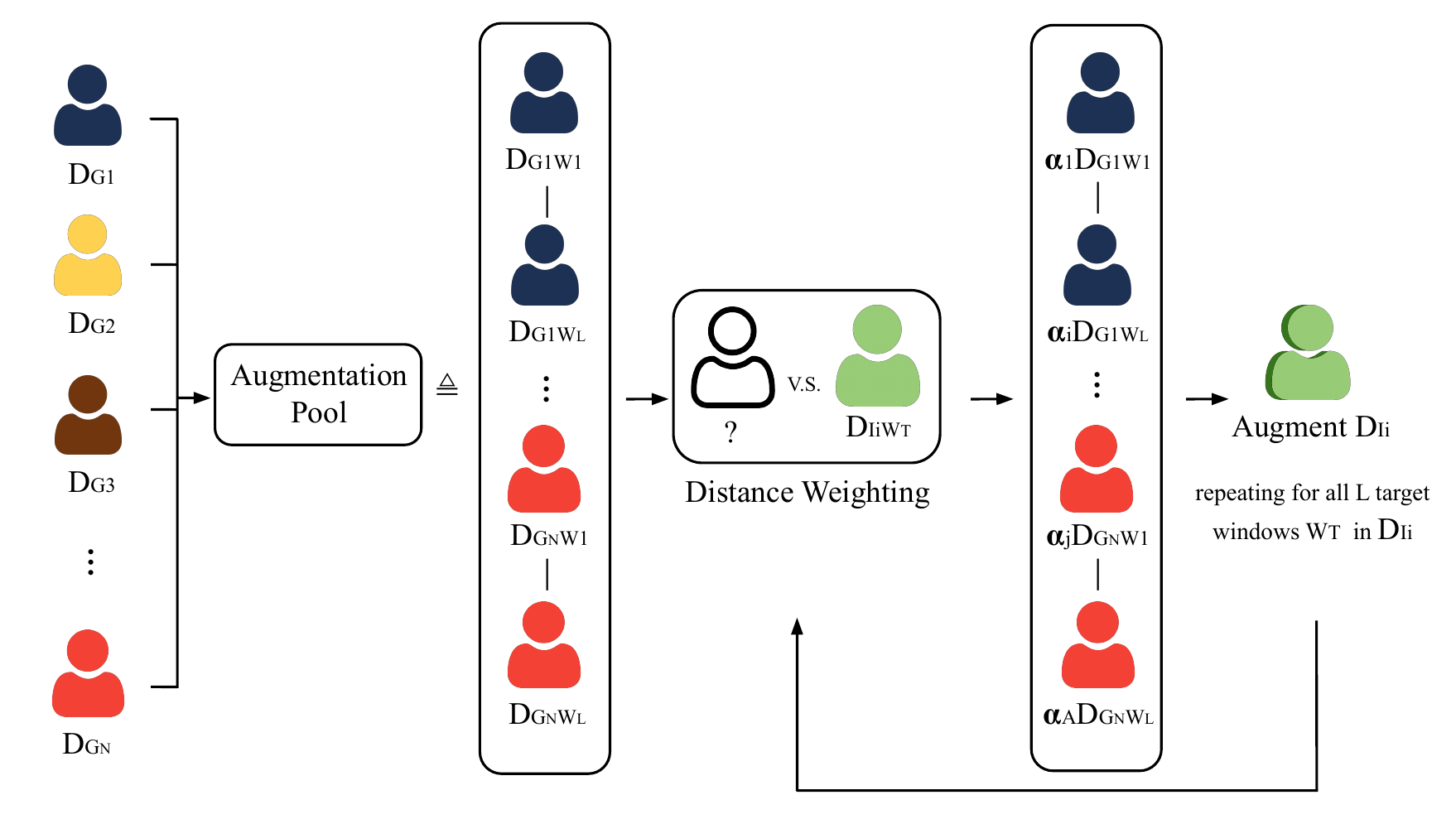}
    \end{center}
  \caption{ DWA generates an dataset $D_{I_i{aug}}$ from the original dataset $D_{I_i}$, by generating an augmentation pool, and selecting most similar samples from that pool, for target segment $W_T$. }
  \label{fig:augmentation}
\end{figure}

As illustrated in Figure \ref{fig:augmentation}, DWA begins by constructing an augmentation pool $AP$ from $D_G$. $AP$ comprises all samples (i.e., segments)  $S$,  from all individuals in the training and development sets. ``Training and development sets'' here refer to $D_G$, and not to the training and development sets \textit{Train\_I} and \textit{Devel\_I} in $D_{I_i}$ (see table \ref{tab:Table Dataset}). DWA then iterates through all segments in the target $D_{I_i}$. For each target segment $W^T$, DWA computes the distance to $W^T$ from all segments in $AP$. The nearest $n$ segments, measured via a defined distance metric (cf. section \label{distanc-metrics}), are selected, i.e given a weight of 1, and appended to $D_{I_i}$. All other segments in the augmentation pool are momentarily given a weight of 0, till the next iteration. The result is $D_{I_i{aug}}$, an augmented dataset, for each individual in the test set. DWA then returns $D_{I_{i}aug}$, which consists of the union of $D_{I_i}$ with these $n$ samples for all segments in $D_{I_i}$. $D_{I_{i}aug}$ is then utilised for fine-tuning a generic model $M_{G}$, to obtain a personalised model $M_{I_i}$ for the target individual in question. All individuals in the testing set make use of the same augmentation pool, from which we draw samples with replacement.   

The choice of the distance metric influences the calculated distance values and rankings of the $n$ nearest segments. The subsequent section provides a detailed description of these metrics.

\subsection{Distance Metrics}\label{distanc-metrics}

We employed three distance metrics during augmentation, to compute the similarity between two time segments of equal length: (1) centroid L2 distance, (2) centroid dot product, and (3) cosine distance. These were chosen as well studied distance measures suitable to our problem statement. For a time segment $W^T$, we have a set of features $\{W^T_j\}_{j=1}^{winlen}$, where the number of features depends on the segment length. For instance,  ${winlen}=10$ implies that 10 timestamps are grouped together within each segment, with 1 feature vector for each timestamp. By $\overline{W^T}$, we denote the average of all ${winlen}$ feature vectors belonging to segment $W^T$. This is equivalent to computing the centroid feature vector, and it is utilised in Algorithm \ref{alg:algo1}. For two time segments $W^1$ and $W^2$, our three distance metrics can be defined as follows: 

\subsubsection{Centroid L2 Distance}
Centroid L2 Distance calculates segment similarity by measuring the L2 distance between the mean vectors of two given segments. The mean vector within a given segment represents the centroid, or average feature over that time frame. 
L2 distance, also denoted as euclidean distance, has been proven effective in the literature for measuring similarity between feature vectors in various domains, such as face recognition \cite{malkauthekar2013analysis}, speaker voice similarity  \cite{singh2019speaker,san2017euclidean}, and text similarity \cite{vijaymeena2016survey,huang2008similarity} systems. A smaller Euclidean distance between two vectors suggests a higher degree of similarity, with a value of zero indicating that the vectors are identical.  
\begin{equation}  \lVert \overline{W^1} - \overline{W^2} \rVert _2 \end{equation} 

\subsubsection{Centroid Dot Product (centroid DP) Distance} 
Similar to the Centroid L2 calculation, we utilise Centroid Dot Product distance to compute the distance between the centroids of two respective segments. However, as dot product is inherently a measure of similarity, we take the negative of the dot product as a measure of distance. The effectiveness of dot product as a measure of similarity is time-tested, not the least by virtue of its widespread usage in transformer models \cite{vaswani2017attention}.

\begin{equation} - \overline{W^1} \cdot \overline{W^2} \end{equation}

\subsubsection{Cosine Distance}
Cosine distance is a robust and efficient metric that excels in high-dimensional applications such as text mining, information retrieval, and computer vision \cite{vijaymeena2016survey}. Originating from the widely-used cosine distance measure in vector space \cite{li2013distance,al2017toward}, it quantifies the distance between pair-wise vectors or segments. Cosine distance is scale-invariant, focusing on the orientation of vectors rather than their magnitude, ensuring its effectiveness for computing distance for sparse and high-dimensional feature vectors in \textsc{Ulm-TSST} dataset. Cosine distance has a bounded range from 0 to 1, with 0 indicating identical vectors.

\begin{equation} \displaystyle \sum_j \left( 1- \frac{W^1_j \cdot W^2_j}{\lVert W^1_j \rVert _2 - \lVert W^2_j \rVert _2} \right) \end{equation} 

\section{EXPERIMENTAL SETUP}
In this section, we describe the experimental setup including the baseline we compare our approach to, dataset, loss function,  evaluation metric, and experimental details.  
\subsection{Baseline Models}
The baseline work from \cite{christ2023muse}, much like our work, uses a two-stage transfer learning approach. In the first stage, it trains a generic model $M_G$, using the generic dataset $D_G$. It then copies the generic model multiple times for each individual in the testing set, and trains separate personalised models using each $D_{I_i}$. Similarly to our approach, the training and development parts, \textit{Train\_I} and \textit{Devel\_I}, of the \textit{Test} individuals, are used to personalise the respective generic models. Our approach seeks to build upon this baseline using DWA, whereby each \textit{Train\_${I_i}$} is augmented with similar samples from $D_G$.
  
\subsection{Dataset}
We utilise the \textsc{Ulm-TSST} dataset from the MuSe 2023 Personalisation sub-challenge \cite{christ2023muse}. This dataset comprises video, audio, written transcripts, and physiological signals obtained from approximately 5 minutes of solo conversations with 69 different individuals. In each solo conversation, a single individual stands in front of a camera and speaks continuously in the German language. The dataset labels consist of continuous valence and arousal labels for small incremental timestamps through the length of each conversation. It is vital to note that the dataset is divided into training, development, and testing sets based on the individuals themselves, rather than time segments. That is, 41 individuals are assigned to the training set, 14 individuals to the development set, and 14 individuals to the testing set. In this paper we refer to the 55 training and development individuals as $D_G$, while the 14 testing individuals are represented as $D_I = \{D_{I_i}\}_{i=1}^{14}$. 

Our methodological objective is to accurately predict the arousal and valence values on the 14 individuals for the testing set. Notably, while labels are available for the entire 5 minutes of conversation for both the training and development individuals, we are only provided labels for 2 minutes of conversation of each testing individual. That is, for each testing individual, the first labeled minute is for personalised training and the second minute is for personalised development/validation. This allows us to conduct personalised fine-tuning for each testing individual $D_{I_i}$ after training a generic model on $D_G$. A summary of the available labels for the train-test split is provided in table \ref{tab:Table Dataset}. Notably, the testing portion of the testing set lacks labels, and our objective in the MuSe-Personalisation challenge is to accurately predict the valence and arousal labels for this unlabeled part of the testing set.

\textbf{Features. }  For the different modalities, pre-extracted feature sets were made available by \cite{christ2023muse}. These are summarised in table \ref{tab:feature_summary}. Among the video features, \textsc{Fau} most directly encodes facial action units \cite{zhi2020comprehensive}, while  \textsc{FaceNet512} features are associated with deep facial features \cite{schroff2015facenet}. While \textsc{ViT} \cite{dosovitskiy2020image} features encode information about visual semantics, these semantics may not be facial information. This is because the \textsc{ViT} model is pre-trained on Imagenet \cite{deng2009imagenet} using DINO \cite{caron2021emerging} self-supervision. Among the audio features, \textsc{eGeMAPS}  \cite{eyben2015geneva} are based on a set of interpretable speech parameters, which can be computed in closed form from the data. Each \textsc{eGeMAPS} dimension corresponds to a parameter such as pitch, shimmer, loudness, or alpha-ratio. \textsc{DeepSpectrum} and \textsc{Wav2Vec} features are extracted via neural network. However, \textsc{DeepSpectrum} \cite{amiriparian2017snore} features are extracted from spectrogram images using a convolutional neural network (CNN), while \textsc{Wav2Vec} \cite{wagner2023dawn} features are extracted from audio waveforms using a combination of CNNs and transformers.  \textsc{Bert-4} \cite{devlin2018bert} features are extracted using a BERT transformer, which has been pretrained on the German language. The  \textsc{Biosignals} features consist of machine-recorded readings of electrocardiograms (ECG), Respiration (RESP) and heart rate beats per minute (BPM).

\begin{table}[]
\caption{Understanding the features in our dataset.}
\label{tab:feature_summary}
\begin{center}  
\resizebox{1.0\columnwidth}{!}{%
\begin{tabular}{@{}c|lcl@{}}
\toprule
\textbf{Modality} & \textbf{Feature} & \textbf{Dim.} & \textbf{Information} \\  
 \midrule
\multirow{4}{*}{\textbf{Video}} & \textsc{FaceNet512} \cite{schroff2015facenet} & 512 &  Facial structure \\
& \textsc{Fau}\footnotemark ~\cite{zhi2020comprehensive}& 20 & Facial actions \\
& \multirow{2}{*}{\textsc{ViT} \cite{caron2021emerging}} & \multirow{2}{*}{384} & Visual semantics \\& & & (not necessarily facial) \\
\midrule
\multirow{4}{*}{\textbf{Audio}}& \multirow{2}{*}{\textsc{eGeMAPS}\cite{eyben2015geneva}} & \multirow{2}{*}{88}& Speech features  \\
& & & (frequency, amplitude, etc.) \\
& \textsc{DeepSpectrum} \cite{amiriparian2017snore} & 1024 & Spectrogram image features \\
& \textsc{Wav2Vec} \cite{wagner2023dawn}& 1024 & Audio wave features\\
\midrule
\textbf{Text} & \textsc{Bert-4}\footnotemark ~\cite{devlin2018bert}& 768 & Transcription semantics \\
\midrule
\multirow{2}{*}{\textbf{Physiological}} & \multirow{2}{*}{\textsc{Biosignals}} & \multirow{2}{*}{3} & Heart rate/electrical\\
& & &  activity, breathing \\
  \bottomrule                 
\end{tabular}
}
\end{center}
\end{table}

\begin{table}[]
\captionsetup{width=0.84\columnwidth}
\caption{Understanding the structure of the dataset labels provided. The number indicate how many individuals belong to the respective tasks.}
\label{tab:Table Dataset}
\begin{center}  
\resizebox{0.84\columnwidth}{!}{%
\begin{tabular}{@{}cc|ccc@{}}
\toprule
 \multicolumn{2}{c|}{\textbf{$D_G$}} & \multicolumn{3}{c}{\textbf{$\{D_{I_i}\}$}}  \\
 \midrule
\multirow{2}{*}{\textbf{Train\_G (41)}} & \multirow{2}{*}{\textbf{Devel\_G (14)}} & \multicolumn{3}{c}{\textbf{Test (14)}}  \\  
\cmidrule(l){3-5} &  & \textit{Train\_I} & \textit{Devel\_I} & Test\\ \midrule 
  \cmark &   \cmark & \cmark &  \cmark  & \xmark\\
  \bottomrule                 
\end{tabular}
}
\end{center}
\end{table}

\footnotetext[1]{\scriptsize \texttt{https://py-feat.org/pages/au\_reference.html}}

\footnotetext{\scriptsize \texttt{https://huggingface.co/dbmdz/bert-base-german-cased}}

\subsection{Loss Function and Evaluation Metric}\label{loss-function}
In line with \cite{christ2023muse}, we adopt the Concordance Correlation Coefficient (CCC) as the loss function to train our model. It is typically used to assess how well two sets of observations agree with each other in terms of both their mean values and their dispersion, i.e. accuracy and precision. As a product of the Pearson Correlation Coefficient (PCC) and the Bias Correction Factor (BCF), the resulting value of CCC ranges between $-1$ and $1$, where a value of $1$ indicates perfect agreement, $0$ indicates no agreement, and $-1$ indicates perfect disagreement. The CCC loss is formulated as follows:

\begin{equation}
\centering
Loss = 1 - CCC
\end{equation}
\begin{equation}
\centering
CCC = \frac{2 \rho \sigma_x \sigma_y}{\sigma_x^2 + \sigma_y^2 + (\mu_x - \mu_y)^2}
\end{equation}

 where $\mu_{\hat{y}}$ and $\mu_{y}$ represent the mean of the prediction $\hat{y}$ and the label $y$, respectively. Similarly, $\sigma_{\hat{y}}$ and $\sigma_{y}$ correspond to their standard deviations. $\rho$ denotes the PCC between $\hat{y}$ and $y$. It measures the strength and direction of the linear relationship between $X$ and $Y$. A value of $\rho$ close to $1$ indicates a strong positive linear relationship, while a value close to $-1$ indicates a strong negative linear relationship. A value of $0$ indicates no linear relationship. In addition, $\frac{2 \sigma_x \sigma_y}{\sigma_x^2 + \sigma_y^2 + (\mu_x - \mu_y)^2}$ denotes the BCF. It corrects for any deviation of the means of $X$ and $Y$ from a perfect linear relationship. It ensures that the CCC accounts for the accuracy component as well. BCF ranges between $0$ and $1$, where a value of $1$ indicates perfect accuracy.

\begin{table}[!ht]
\caption{Performance of Generic models on $D_I$, as a benchmark prior to personalisation. \hlc[yellow]{Yellow}: maximum and minimum scores. \textbf{Bold:} Best performing feature in the specific modality. \underline{\textbf{Bold:}} Best performance overall the features. }
\label{tab: M_G training}
\begin{center}
\resizebox{1.0\linewidth}{!}{%
\begin{tabular}{@{}ccccccc@{}}
\toprule 
\multirow{2}{*}{\textbf{Feature}}  & \multicolumn{2}{c}{\textbf{Arousal}} & \multicolumn{2}{c}{\textbf{Valence}} & \textbf{Combined} \\ 
\cmidrule(l){2-6}  
 &  \textit{Devel\_I} & Test & \textit{Devel\_I} & Test  & Test \\ 
 \midrule 
\textsc{FaceNet512} & 0.43 & 0.49 & 0.76 & \underline{\textbf{0.56}} & \hlc[yellow]{0.52}\\
\textsc{Fau} & 0.21 & -0.00 & 0.52 & 0.47 & 0.23\\
\textsc{ViT} & 0.61 & \textbf{0.51} & 0.40 & 0.40 & 0.45\\
\midrule 
\textsc{eGeMAPS} & 0.58 & \underline{\textbf{0.56}} & 0.58 & \textbf{0.49} & \hlc[yellow]{0.52}\\
\textsc{DeepSpectrum} & 0.54 & 0.42 & 0.57 & 0.45 & 0.44\\ 
\textsc{Wav2Vec}  & 0.52 & 0.54 & 0.46 & 0.42 & 0.48\\  
\textsc{Bert-4} & 0.51 & 0.49 & 0.34 & 0.25 & 0.37\\
\midrule 
\textsc{Biosignals} & -0.14 & -0.14 & 0.23 & 0.19 & \hlc[yellow]{0.03}\\  \bottomrule
\end{tabular} 
}
\end{center}
\end{table}

\subsection{Experimental Details} 

\textbf{Baseline Models.} The baseline work of \cite{christ2023muse}, while insightful, only reports results for 6 audio and video features, namely, Fau, FaceNet512, ViT, eGeMAPS, DeepSpectrum, and Wav2Vec. We conduct all our experiments on these features, but also include results for  the \textsc{Bert-4}, and \textsc{Biosignals} features. 
We also employ the same generic model from the baseline methodology for consistent comparison.

\textbf{Hyperparameters.} DWA was used to augment each individual dataset and obtain $D_{I_{\text{aug}}}$.  For fair comparison with the baseline, we set consistent ranges for any tuned hyerparameters such as learning rates, number of RNN layers, and model dimensions, during hyperparameter search. All  other hyperparameters, including early stopping patience, segment length, and hop length, are kept unchanged from the baseline method in \cite{christ2023muse}.  However, we perform the hyperparameter search for all 8 features in the dataset, expanding upon the 6 reported features of the \cite{christ2023muse}. 

Our experiments sought to understand the impact of different choices of distance metrics and numbers of augmentation samples per segment. We trained models using data augmented with one of the three distance metrics: centroid DP, cosine, and centroid L2. We also explored varying values for number of augmentation samples per segment $n$, specifically $n=1$ up to $3$.

\begin{table}[]
\caption{ Performance of baseline personalised models on $D_I$, to highlight the effect of personalisation. \hlc[yellow]{Yellow}: greatest increases due to personalisation w.r.t generic models. \textbf{Bold:} Best performing feature in the specific modality. \underline{\textbf{Bold:}} Best performance overall the features.}
\label{tab: M_I results}
\begin{center}
\resizebox{1.0\linewidth}{!}{%
\begin{tabular}{@{}cccccc@{}}
\toprule 
\multirow{2}{*}{\textbf{Feature}} &  \multicolumn{2}{c}{\textbf{Arousal}} & \multicolumn{2}{c}{\textbf{Valence}} & \textbf{Combined} \\ 
\cmidrule(l){2-6}  
  &  \textit{Devel\_I} & Test & \textit{Devel\_I} & Test & Test \\ 
 \midrule 
\textsc{FaceNet512} & 0.88 & 0.68 & 0.89 & 0.69 & 0.69 \\ 
\textsc{Fau} & 0.88 & \underline{\textbf{0.79}} & 0.90 & \textbf{0.78} & \hlc[yellow]{0.79} \\
\textsc{ViT} & 0.85 & 0.64 & 0.67 & 0.55 & 0.60 \\
\midrule 
\textsc{eGeMAPS} & 0.70 & 0.53 & 0.68 & 0.59 & 0.56 \\ 
\textsc{DeepSpectrum} & 0.90 & 0.55 & 0.72 & 0.56 & 0.55 \\
\textsc{Wav2Vec} & 0.89 & 0.53 & 0.90 & \underline{\textbf{0.80}} & 0.66 \\  
\textsc{Bert-4} & 0.90 & \textbf{0.76} & 0.87 & 0.60 & 0.68 \\
\midrule 
\textsc{Biosignals} & 0.78 & 0.61 & 0.45 & 0.35 & \hlc[yellow]{0.48} 
\\ \bottomrule

\end{tabular} 
}
\end{center}
\end{table}

\begin{table*}[!h]
\caption{Performance of DWA with exploration of $n$ and distance metrics. \hlc[green]{Green}: DWA outperforms the baseline on \textit{Devel\_I}. \hlc[cyan]{Blue}: DWA outperforms baseline on Test set. \hlc[yellow]{Yellow}: used in place of blue if improvement is $> 0.1$ }
\label{tab: DWA results}
\begin{center}
\begin{tabular}{@{}cccccc|cccc|c@{}}
\toprule 
\multirow{2}{*}{\textbf{Feature}} & \multirow{2}{*}{\textbf{Distance Metric}}  & \multicolumn{4}{c|}{\textbf{Arousal}} & \multicolumn{4}{c|}{\textbf{Valence}} & \textbf{Combined } \\ 
\cmidrule(l){3-11}  
 & &  n=1 &  n=2 &  n=3  & Test & n=1 &  n=2 &  n=3 & Test  & Test \\ 
 \midrule 
\multirow{3}{*}{\textsc{FaceNet512}}  & Centroid\_DP & 0.75 & 0.81 & 0.71 & \hlc[cyan]{0.68} & \hlc[lime]{0.92} & \hlc[lime]{0.92} & \hlc[lime]{0.91} & \hlc[cyan]{0.78} & 0.72 \\
  & Centroid\_L2 & 0.83 & 0.68 & 0.82 & \hlc[cyan]{0.69} & \hlc[lime]{0.94} & \hlc[lime]{0.95} & \hlc[lime]{0.94} & \hlc[cyan]{0.79} & 0.74 \\
  & Cosine & 0.83 & 0.83 & 0.72 & 0.65 & \hlc[lime]{0.90} & \hlc[lime]{0.92} & \hlc[lime]{0.93} & \hlc[yellow]{0.83} & \hlc[cyan]{0.74} \\
\midrule 
\multirow{3}{*}{\textsc{Fau}} & Centroid\_DP & 0.72 & 0.76 & 0.81 & 0.60 & 0.76 & 0.77 & 0.80 & 0.64 & 0.67 \\
& Centroid\_L2 & 0.79 & 0.73 & 0.71 & 0.55 & 0.90 & 0.74 & 0.66 & 0.83 & 0.69 \\
  & Cosine & 0.74 & 0.83 & 0.82 & 0.71 & 0.77 & 0.79 & 0.72 & 0.65 & 0.74 \\
\midrule 
\multirow{3}{*}{\textsc{ViT}} & Centroid\_DP & 0.84 & 0.84 & \hlc[lime]{0.92} & \hlc[yellow]{0.71} & \hlc[lime]{0.78} & 0.54 & 0.54 & \hlc[yellow]{0.65} & \hlc[yellow]{0.67} \\
& Centroid\_L2 & \hlc[lime]{0.89} & 0.78 & 0.85 & \hlc[cyan]{0.64} & 0.66 & \hlc[lime]{0.69} & 0.52 & 0.54 & \hlc[cyan]{0.59} \\
  & Cosine & 0.85 & \hlc[lime]{0.87} & \hlc[lime]{0.93} & \hlc[cyan]{0.70} & 0.51 & 0.59 & 0.54 & 0.52 & \hlc[cyan]{0.61} \\
\midrule 
\multirow{3}{*}{\textsc{eGeMAPS}} & Centroid\_DP & \hlc[lime]{0.75} & 0.69 & \hlc[lime]{0.75} & \hlc[yellow]{0.71} & \hlc[lime]{0.68} & \hlc[lime]{0.81} & \hlc[lime]{0.81} & \hlc[cyan]{0.68} & \hlc[yellow]{0.70}  \\
& Centroid\_L2 & \hlc[lime]{0.74} & 0.66 & \hlc[lime]{0.74} & 0.09 & \hlc[lime]{0.73} & 0.59 & \hlc[lime]{0.69} & 0.46 & 0.27 \\
  & Cosine & \hlc[lime]{0.82} & \hlc[lime]{0.82} & 0.62 & 0.09 & 0.65 & 0.60 & 0.64 & 0.59 & 0.34 \\
\midrule 
\multirow{3}{*}{\textsc{DeepSpectrum}} & Centroid\_DP & 0.90 & 0.68 & 09.73 & 0.59 & \hlc[lime]{0.78} & \hlc[lime]{0.76} & 0.67 & 0.38 & 0.49 \\
& Centroid\_L2 & 0.85 & 0.69 & 0.66 & \hlc[cyan]{0.58} & \hlc[lime]{0.84} & \hlc[lime]{0.78} & \hlc[lime]{0.76} & \hlc[cyan]{0.63} & \hlc[cyan]{0.60} \\
  & Cosine & 0.79 & 0.79 & 0.60 & 0.28 & \hlc[lime]{0.78} & \hlc[lime]{0.80} & \hlc[lime]{0.75} &  \hlc[yellow]{0.72} & 0.50 \\
\midrule 
\multirow{3}{*}{\textsc{Wav2Vec}} & Centroid\_DP & 0.80 & 0.81 & 0.70 & 0.52 & 0.72 & 0.58 & 0.59 & 0.70 & 0.61 \\
& Centroid\_L2 & 0.74 & 0.76 & 0.60 & 0.54 & 0.50 & 0.72 & 0.60 & 0.65 & 0.59 \\
  & Cosine & 0.73 & 0.80 & 0.83 & 0.60 & 0.84 & 0.70 & 0/64 & 0.68 & 0.64 \\
\midrule   
\multirow{3}{*}{\textsc{Bert-4}} &  Centroid\_DP & 0.82 & 0.79 & 0.88 & 0.67 & 0.82 & 0.59 & 0.78 & 0.60 & 0.63 \\
& Centroid\_L2 & 0.85 & 0.89 & 0.77 & 0.57 & 0.78 & 0.53 & 0.76 & 0.46 & 0.51 \\
  & Cosine & 0.90 & 0.82 & \hlc[lime]{0.91} & 0.73 & 0.80 & 0.82 & 0.75 & 0.47 & 0.60 \\
\midrule 
\multirow{3}{*}{\textsc{Biosignals}} & Centroid\_DP & 0.28 & 0.19 & 0.50 & 0.50 & 0.42 & 0.26 & 0.29 & 0.14 & 0.32 \\ 
& Centroid\_L2 & 0.35 & 0.12 & 0.09 & 0.20 & \hlc[lime]{0.51} & \hlc[lime]{0.77} & \hlc[lime]{0.64} & \hlc[yellow]{0.69} & 0.44 \\
  & Cosine & 0.67 & 0.48 & 0.53 & 0.56 & \hlc[lime]{0.74} & \hlc[lime]{0.67} & \hlc[lime]{0.61} & \hlc[cyan]{0.45} & \hlc[cyan]{0.50} \\\bottomrule

\end{tabular} 
\end{center}
\end{table*}

%
\begin{table*}[!ht]
\caption{Results from late fusion, for features where DWA showed improvements in ablation. \hlc[yellow]{Yellow}: improvements over the baseline in \hlc[cyan]{Blue}. }
\label{tab: fusion comparisons}
\begin{center}
\resizebox{0.65\linewidth}{!}{%
\begin{tabular}{@{}cccccc@{}}
\toprule 
\multirow{2}{*}{\textbf{Feature}}  & \multicolumn{2}{c}{\textbf{Arousal}} & \multicolumn{2}{c}{\textbf{Valence}} & \textbf{Combined} \\ 
\cmidrule(l){2-6}  
 & \textit{Devel\_I}  & Test & \textit{Devel\_I} & Test  & Test \\ 
\midrule 

\multicolumn{6}{c}{\textbf{Baseline(s)}}\\ 
\midrule
\textsc{FaceNet512} + \textsc{ViT} & 0.90 & \hlc[cyan]{0.74} & 0.82 & \hlc[cyan]{0.65} & \hlc[cyan]{0.70} \\
Reproduced (\textsc{FaceNet512} + \textsc{eGeMAPS})& 0.85 & 0.67 & 0.90 & \hlc[cyan]{0.76} & \hlc[cyan]{0.72} \\ 
Reported (\textsc{FaceNet512} + \textsc{eGeMAPS}) & 0.92  & 0.75  & 0.86  & 0.78  & 0.76  \\ 

\midrule 
\multicolumn{6}{c}{\textbf{Late Fusion, Proportional Weighting}}\\
\midrule 
\textsc{ViT}+ \textsc{eGeMAPS} + DWA& 0.92 & 0.44 & 0.86 & 0.77 & 0.61 \\
\textsc{FaceNet512} + \textsc{eGeMAPS}+DWA& 0.89 & 0.61 & 0.95 & \hlc[yellow]{0.81} & 0.71 \\
\textsc{ViT}+\textsc{Bert-4}+DWA& 0.93 & 0.76 & 0.88 & 0.68 & 0.72  \\ 
\textsc{ViT}+\textsc{DeepSpectrum}+DWA& 0.94 & 0.71 & 0.88 & 0.74 & 0.73 \\
\midrule
Facenet+\textsc{ViT}+DWA & 0.92 & \hlc[yellow]{0.76} & 0.94 & \hlc[yellow]{0.80} & \hlc[yellow]{0.78} \\

\bottomrule

\end{tabular} 
}
\end{center}
\end{table*}

\section{RESULTS AND DISCUSSION}
We organise our discussion into 2 sections. Firstly, we seek to highlight the experimental effects of personalisation. To do this, we reproduce the generic model and personalised model results of the baseline paper on our dataset, and compare their respective performances on the testing set (section \ref{res:pers-effect}). This is particularly important in the context of personalisation, as the generic model results on the testing set were not provided in the baseline work \cite{christ2023muse}. Secondly, we perform a series of experiments to evaluate the performance of the DWA personalisation approach (section \ref{res:dwa}). 
Our objective is to show that augmentation with individual-specific samples from our augmentation pool can enhance the performance of personalised model training on valence and arousal prediction for each testing individual.

\subsection{The Effect of Personalisation} \label{res:pers-effect}
\textbf{Generic Model Performance Evaluation.} The findings presented in table \ref{tab: M_G training} serve to demonstrate the need for personalisation, by demonstrating scores of the generic model $M_G$, obtained without the use of personalisation, in order to have a reference for any effects that may result from a personalised training step. We observe that nearly all models have combined CCC scores below 0.53. The highest performing features are \textsc{FaceNet512} and \textsc{eGeMAPS}, with combined testing scores of 0.52. In particular, \textsc{eGeMAPS} scored highest on arousal with a testing score of 0.56. While  \textsc{FaceNet512} had highest such score of 0.56 on valence. This suggests that without personalisation, a model trained on a general dataset will be less effective on a given individual. The lowest testing CCC scores are obtained by \textsc{Fau},  and \textsc{Biosignals} features, which score 0.23 and 0.03 respectively. The \textsc{Biosignals} feature scores lowest in both cases of arousal and valence, obtaining -0.14, and 0.19 respectively. For each timestamp, the \textsc{Biosignals} feature vector comprises 3 measurements: 1 for ECG, 1 for resps, and 1 for BPM. We hypothesise that the low dimensionality of the \textsc{Biosignals} feature may be one additional reason for its extraordinarily low performance. However, all in all, the generic model results serve as a backdrop to compare personalised model scores with and without DWA. 

\textbf{Baseline Personalisation Performance Evaluation.} The development and testing set CCC scores obtained from personalised training using the baseline approach \cite{christ2023muse} across all features, are presented in table \ref{tab: M_I results}. The \textsc{Fau} feature, which previously scored 0.23 (combined CCC) in our reproduced results, without personalisation, now obtains the highest CCC across all features, of 0.79, owed to respective arousal and valence scores of 0.79 and 0.78 respectively. The \textsc{Biosignals}  feature still has the lowest score, at 0.48, but dramatically outperforms its corresponding generic model score of 0.03. \textsc{Biosignals} has the 5th highest score in arousal of 0.61, but scores only 0.35 in valence performance. In our experiments, while \textsc{FaceNet512} and \textsc{eGeMAPS} improved from a testing CCC of 0.52 to 0.69, and 0.56 respectively, they are by no means our best performing respective audio and video features. These were instead \textsc{Fau} and \textsc{Wav2Vec}, which improved from 0.23 to 0.79, and from 0.48 to 0.66 respectively, thanks to personalisation. In the case of \textsc{Wav2Vec}, which was the best performing audio feature, it scores 0.80 in valence, making it the highest performing feature in valence. However, it scored 0.53 on arousal.

\subsection{Performance Evaluation of DWA}\label{res:dwa}
 The 8 different feature types utilised in our experiments, each capture different characteristics of a target individual. For example, \textsc{FaceNet512} features capture facial characteristics, while \textsc{Fau} features capture facial actions, and \textsc{eGeMAPS} capture variations in prosody. Yet fundamentally, DWA augments a given segment with other segments from an augmentation pool, based on measurements of distance. Therefore, the efficacy of DWA for a given feature depends on the meaning of the distance metric in that feature space. For example, \textsc{FaceNet512} features encode a compact representation of facial features in the Euclidean space by which distances correspond to a measure of face similarity, thus DWA may result in augmentation with similar faces. Thus, in order to better understand the performance of DWA, we present our experiments across all features in table \ref{tab: DWA results}, recognising that DWA may be more suitable for some features than others. Table \ref{tab: DWA results} also captures the effects of alternative distance metrics, the Centroid DP, Centroid L2, and Cosine metrics. We present varying values of number of augmentation samples per segment between $n=1$ and $n=3$ as well.

\textbf{Arousal vs. Valence.} A key observation from  table \ref{tab: DWA results}, is that DWA may result in improvement in arousal or valence tasks independently. Our experiments suggest that an improvement in valence prediction due to DWA, may not guarantee an improvement in arousal prediction, and vice versa. However, for feature-task combinations where DWA does result in improvement, such improvements are generally seen among all distance metrics, and for most values of $n$, as can be observed by the neat patterns of green highlighting seen in the table. For instance, while the \textsc{DeepSpectrum} feature improves on valence to a score of 0.72, it fails to improve on arousal, peaking at 0.59. On the other hand, the \textsc{eGemMAPS} feature is able to improve on both arousal and valence, reaching scores of 0.71 and 0.68 respectively.

\textbf{Video \& Physiological Features.} Among the video features, only \textsc{FaceNet512} and \textsc{ViT} features seem to benefit from DWA, with the latter seeing the most improvement compared to the baseline model. For \textsc{FaceNet512}, the combined testing CCC peaks at 0.74 with DWA, compared to 0.69 without DWA (baseline model). Much of this improvement is due to the valence prediction, which increases from a testing CCC of 0.69 to 0.83. Arousal prediction does not see similar improvement for  \textsc{FaceNet512}. With \textsc{ViT} features, the final combined testing CCC increases from 0.60 to 0.67. This is due to significant improvements in both arousal prediction from 0.64 to 0.71, and in valence prediction from 0.55 to 0.65 with respect to the baseline.  The \textsc{Fau} features, which had the highest performance without DWA, saw a drop in performance with DWA. This suggests that DWA can only be used as a tool to improve the performance of features which are lagging behind on specific tasks. Indeed, this hypothesis is further supported by the improvement in \textsc{Biosignals} features testing CCC. With the use of DWA a compelling performance increase can be seen from 0.35 to 0.69 for the arousal prediction task. 

\textbf{Audio Features.} Results with audio features are also presented in table \ref{tab: DWA results}. The performance increase due to DWA is most noticeable using the \textsc{eGeMAPS} feature. For the \textsc{eGeMAPS}, DWA brings about an increase in performance from 0.56 to 0.70. This is due to corresponding improvements in testing arousal and valence CCC, which increased from 0.50 and 0.59 respectively, to 0.71 and 0.68 respectively. This combined increase in \textsc{eGeMAPS} performance with to DWA is not only the highest among the audio features, but among all features which were tested. DWA also results in an improved performance in the \textsc{DeepSpectrum} feature from 0.56 to 0.63. However, the improvement is largely limited to the valence prediction task. As a result, since the arousal prediction of \textsc{DeepSpectrum} does not appear to benefit from DWA, the combined CCC score only improves from 0.55 to 0.60.
 
\textbf{Number of Augmentations per segment.} We ran experiments on the value of $n$ to determine the effect of the number of augmentation samples on model performance. What we find, as per table \ref{tab: DWA results}, is that the ideal number of augmentation samples varies depending on the feature and distance metric. For example, consider the case of valence prediction using \textsc{DeepSpectrum} features and cosine distance. The performance on \textit{Devel\_I} increases as n is increasing from 1 to 2, and decreases when n is further increased to 3. This suggests a diminishing return pattern Such is also the case for \textsc{Biosignals} feature. On the other hand, performance on \textsc{FaceNet512}, with cosine feature, seems to be steadily increasing with n. This suggests that the value of n must be tuned uniquely as a hyperparamter for each feature-distance-metric combination.

\textbf{Late Fusion.} Finally, we adopt the same late fusion strategy deployed in our baseline, for comparison purposes. For late fusion, we utilise proportional weighting, whereby two sets of predictions are weighted based on their performance on the development set. For fair comparison, we include the reported results for late fusion from the baseline, of \textsc{FaceNet512} and \textsc{eGeMAPS} features. However, in reproducing these results, we utilise consistent hyperparameter search between all methods, causing the reproduced late fusion score to be different from the reported baseline. The results are presented in table \ref{tab: fusion comparisons}. 

In the baseline work, the best performing feature fusion is of \textsc{FaceNet512} and \textsc{eGeMAPS}, reaching a testing CCC of 0.76. Our reproduced score is slightly lower, at 0.72. However, our DWA method results in a score of 0.78 when fusing \textsc{ViT} and \textsc{eGeMAPS} features. This improvement is significant, as without DWA, the late fusion of \textsc{ViT} and \textsc{eGeMAPS} would only result in a CCC testing score of 0.70. As expected, the fusion of \textsc{eGeMAPS} with \textsc{ViT} scores higher than with \textsc{FaceNet512}, as the latter did not see signficant benefit in arousal prediction from DWA, as earlier discussed. However, this but reinforces our hypothesis, that DWA can be quite useful as a boost to personalisation performance for features which are not performing well.


\vspace{0.5cm}
\section{CONCLUSIONS AND FUTURE WORKS}
\vspace{0.4cm}
In this paper, we introduced the Distance Weighting Augmentation (DWA) method for enhancing predictions of valence and arousal, at the data level. In simple terms, DWA expands a personalised dataset with samples from a global dataset that are the most similar. We implement DWA on all 8 features provided in our target dataset, with the best performance observed from a fusion of \textsc{ViT} and  \textsc{eGeMAPS} predictions. Among the 3 distance metrics employed for similarity measurement, all proved to be effective for either valence or arousal, or both, in 4 out of 8 features.  Our findings indicate that DWA is most useful in improving personalisation performance when utilising features which have poor performance for a particular task. Late fusion of the DWA-trained \textsc{ViT} and \textsc{eGeMAPS} models, yields a combined CCC score of 0.78, indicating the effectiveness of DWA, and opening the door for further research into data-level personalisation. 

Future work could expand upon DWA in 3 key ways. Firstly, we can explore further distance metrics to see if they could improve over the 3 used thus far. Secondly, it may be beneficial to explore fractional weighting of similar samples. This could be implemented via sample weighting, whereby we would alternate between normal samples and augmentation samples within a training loop. In effect, this could allow us to weight samples in the interval [0,1] rather than relying solely on binary weights. Thirdly, it is important to investigate the reasons for which DWA brings limited performance to features which already performed well. In effect, modifications to the algorithm itself, may help in expanding upon its applicability to different feature sets. This last point will be easier to explore as we seek to apply DWA to additional datasets for dimensional emotion prediction. 

{\small
\bibliographystyle{ieee}
\bibliography{muse-citation}

\begin{thebibliography}{10}\itemsep=-1pt

\bibitem{abbasi2009towards}
A.~R. Abbasi, N.~V. Afzulpurkar, and T.~Uno.
\newblock Towards emotionally-personalized computing: Dynamic prediction of student mental states from self-manipulatory body movements.
\newblock In {\em 2009 International Conference on Emerging Technologies}, pages 235--240. IEEE, 2009.

\bibitem{ahmed2023systematic}
N.~Ahmed, Z.~Al~Aghbari, and S.~Girija.
\newblock A systematic survey on multimodal emotion recognition using learning algorithms.
\newblock {\em Intelligent Systems with Applications}, 17:200171, 2023.

\bibitem{al2017toward}
F.~S. Al-Anzi and D.~AbuZeina.
\newblock Toward an enhanced arabic text classification using cosine similarity and latent semantic indexing.
\newblock {\em Journal of King Saud University-Computer and Information Sciences}, 29(2):189--195, 2017.

\bibitem{alswaidan2020survey}
N.~Alswaidan and M.~E.~B. Menai.
\newblock A survey of state-of-the-art approaches for emotion recognition in text.
\newblock {\em Knowledge and Information Systems}, 62:2937--2987, 2020.

\bibitem{alyuz2016semi}
N.~Alyuz, E.~Okur, E.~Oktay, U.~Genc, S.~Aslan, S.~E. Mete, B.~Arnrich, and A.~A. Esme.
\newblock Semi-supervised model personalization for improved detection of learner's emotional engagement.
\newblock In {\em Proceedings of the 18th ACM International Conference on Multimodal Interaction}, pages 100--107, 2016.

\bibitem{amiriparian2017snore}
S.~Amiriparian, M.~Gerczuk, S.~Ottl, N.~Cummins, M.~Freitag, S.~Pugachevskiy, A.~Baird, and B.~Schuller.
\newblock Snore sound classification using image-based deep spectrum features.
\newblock 2017.

\bibitem{ayata2020emotion}
D.~Ayata, Y.~Yaslan, and M.~E. Kamasak.
\newblock Emotion recognition from multimodal physiological signals for emotion aware healthcare systems.
\newblock {\em Journal of Medical and Biological Engineering}, 40:149--157, 2020.

\bibitem{barros2019personalized}
P.~Barros, G.~Parisi, and S.~Wermter.
\newblock A personalized affective memory model for improving emotion recognition.
\newblock In {\em International Conference on Machine Learning}, pages 485--494. PMLR, 2019.

\bibitem{barros2022ciao}
P.~Barros and A.~Sciutti.
\newblock Ciao! a contrastive adaptation mechanism for non-universal facial expression recognition.
\newblock In {\em 2022 10th International Conference on Affective Computing and Intelligent Interaction (ACII)}, pages 1--8. IEEE, 2022.

\bibitem{bruna2016emotion}
O.~Bruna, H.~Avetisyan, and J.~Holub.
\newblock Emotion models for textual emotion classification.
\newblock In {\em Journal of physics: conference series}, volume 772, page 012063. IOP Publishing, 2016.

\bibitem{caron2021emerging}
M.~Caron, H.~Touvron, I.~Misra, H.~J{\'e}gou, J.~Mairal, P.~Bojanowski, and A.~Joulin.
\newblock Emerging properties in self-supervised vision transformers.
\newblock In {\em Proceedings of the IEEE/CVF international conference on computer vision}, pages 9650--9660, 2021.

\bibitem{chattopadhyay2012multisource}
R.~Chattopadhyay, Q.~Sun, W.~Fan, I.~Davidson, S.~Panchanathan, and J.~Ye.
\newblock Multisource domain adaptation and its application to early detection of fatigue.
\newblock {\em ACM Transactions on Knowledge Discovery from Data (TKDD)}, 6(4):1--26, 2012.

\bibitem{chithrra2022personalized}
V.~V. Chithrra~Raghuram, H.~Salam, J.~Nasir, B.~Bruno, and O.~Celiktutan.
\newblock Personalized productive engagement recognition in robot-mediated collaborative learning.
\newblock In {\em Proceedings of the 2022 International Conference on Multimodal Interaction}, pages 632--641, 2022.

\bibitem{christ2023muse}
L.~Christ, S.~Amiriparian, A.~Baird, A.~Kathan, N.~M{\"u}ller, S.~Klug, C.~Gagne, P.~Tzirakis, E.-M. Me{\ss}ner, A.~K{\"o}nig, et~al.
\newblock The muse 2023 multimodal sentiment analysis challenge: Mimicked emotions, cross-cultural humour, and personalisation.
\newblock {\em arXiv preprint arXiv:2305.03369}, 2023.

\bibitem{chu2016selective}
W.-S. Chu, F.~De~la Torre, and J.~F. Cohn.
\newblock Selective transfer machine for personalized facial expression analysis.
\newblock {\em IEEE transactions on pattern analysis and machine intelligence}, 39(3):529--545, 2016.

\bibitem{cunningham2019audio}
S.~Cunningham, H.~Ridley, J.~Weinel, and R.~Picking.
\newblock Audio emotion recognition using machine learning to support sound design.
\newblock In {\em Proceedings of the 14th International Audio Mostly Conference: A Journey in Sound}, pages 116--123, 2019.

\bibitem{de2023leveraging}
D.~de~Oliveira, N.~R. Prabhu, and T.~Gerkmann.
\newblock Leveraging semantic information for efficient self-supervised emotion recognition with audio-textual distilled models.
\newblock {\em arXiv preprint arXiv:2305.19184}, 2023.

\bibitem{deng2009imagenet}
J.~Deng, W.~Dong, R.~Socher, L.-J. Li, K.~Li, and L.~Fei-Fei.
\newblock Imagenet: A large-scale hierarchical image database.
\newblock In {\em 2009 IEEE conference on computer vision and pattern recognition}, pages 248--255. Ieee, 2009.

\bibitem{devlin2018bert}
J.~Devlin, M.-W. Chang, K.~Lee, and K.~Toutanova.
\newblock Bert: Pre-training of deep bidirectional transformers for language understanding.
\newblock {\em arXiv preprint arXiv:1810.04805}, 2018.

\bibitem{dosovitskiy2020image}
A.~Dosovitskiy, L.~Beyer, A.~Kolesnikov, D.~Weissenborn, X.~Zhai, T.~Unterthiner, M.~Dehghani, M.~Minderer, G.~Heigold, S.~Gelly, et~al.
\newblock An image is worth 16x16 words: Transformers for image recognition at scale.
\newblock {\em arXiv preprint arXiv:2010.11929}, 2020.

\bibitem{eyben2015geneva}
F.~Eyben, K.~R. Scherer, B.~W. Schuller, J.~Sundberg, E.~Andr{\'e}, C.~Busso, L.~Y. Devillers, J.~Epps, P.~Laukka, S.~S. Narayanan, et~al.
\newblock The geneva minimalistic acoustic parameter set (gemaps) for voice research and affective computing.
\newblock {\em IEEE transactions on affective computing}, 7(2):190--202, 2015.

\bibitem{hasnul2021electrocardiogram}
M.~A. Hasnul, N.~A.~A. Aziz, S.~Alelyani, M.~Mohana, and A.~A. Aziz.
\newblock Electrocardiogram-based emotion recognition systems and their applications in healthcare—a review.
\newblock {\em Sensors}, 21(15):5015, 2021.

\bibitem{hazourli2021multi}
A.~R. Hazourli, A.~Djeghri, H.~Salam, and A.~Othmani.
\newblock Multi-facial patches aggregation network for facial expression recognition and facial regions contributions to emotion display.
\newblock {\em Multimedia Tools and Applications}, 80:13639--13662, 2021.

\bibitem{huang2008similarity}
A.~Huang et~al.
\newblock Similarity measures for text document clustering.
\newblock In {\em Proceedings of the sixth new zealand computer science research student conference (NZCSRSC2008), Christchurch, New Zealand}, volume~4, pages 9--56, 2008.

\bibitem{jaques2016multi}
N.~Jaques, S.~Taylor, E.~Nosakhare, A.~Sano, and R.~Picard.
\newblock Multi-task learning for predicting health, stress, and happiness.
\newblock In {\em NIPS Workshop on Machine Learning for Healthcare}, 2016.

\bibitem{jensen2016personality}
M.~Jensen.
\newblock Personality traits and nonverbal communication patterns.
\newblock {\em Int'l J. Soc. Sci. Stud.}, 4:57, 2016.

\bibitem{kasparova2020inferring}
A.~Kasparova, O.~Celiktutan, and M.~Cukurova.
\newblock Inferring student engagement in collaborative problem solving from visual cues.
\newblock In {\em Companion Publication of the 2020 International Conference on Multimodal Interaction}, pages 177--181, 2020.

\bibitem{kathan2022personalised}
A.~Kathan, S.~Amiriparian, L.~Christ, A.~Triantafyllopoulos, N.~M{\"u}ller, A.~K{\"o}nig, and B.~W. Schuller.
\newblock A personalised approach to audiovisual humour recognition and its individual-level fairness.
\newblock In {\em Proceedings of the 3rd International on Multimodal Sentiment Analysis Workshop and Challenge}, pages 29--36, 2022.

\bibitem{kim2009towards}
J.~Kim, E.~Andr{\'e}, and T.~Vogt.
\newblock Towards user-independent classification of multimodal emotional signals.
\newblock In {\em 2009 3rd International Conference on Affective Computing and Intelligent Interaction and Workshops}, pages 1--7. IEEE, 2009.

\bibitem{kollias2023abaw}
D.~Kollias, P.~Tzirakis, A.~Baird, A.~Cowen, and S.~Zafeiriou.
\newblock Abaw: Valence-arousal estimation, expression recognition, action unit detection \& emotional reaction intensity estimation challenges.
\newblock In {\em Proceedings of the IEEE/CVF Conference on Computer Vision and Pattern Recognition}, pages 5888--5897, 2023.

\bibitem{leyzberg2014personalizing}
D.~Leyzberg, S.~Spaulding, and B.~Scassellati.
\newblock Personalizing robot tutors to individuals' learning differences.
\newblock In {\em Proceedings of the 2014 ACM/IEEE international conference on Human-robot interaction}, pages 423--430, 2014.

\bibitem{li2013distance}
B.~Li and L.~Han.
\newblock Distance weighted cosine similarity measure for text classification.
\newblock In {\em Intelligent Data Engineering and Automated Learning--IDEAL 2013: 14th International Conference, IDEAL 2013, Hefei, China, October 20-23, 2013. Proceedings 14}, pages 611--618. Springer, 2013.

\bibitem{li2023survey}
J.~Li, A.~Waleed, and H.~Salam.
\newblock A survey on personalized affective computing in human-machine interaction.
\newblock {\em arXiv preprint arXiv:2304.00377}, 2023.

\bibitem{li2020personality}
L.~Li, H.~Zhu, S.~Zhao, G.~Ding, and W.~Lin.
\newblock Personality-assisted multi-task learning for generic and personalized image aesthetics assessment.
\newblock {\em IEEE Transactions on Image Processing}, 29:3898--3910, 2020.

\bibitem{liang2020pose}
G.~Liang, S.~Wang, and C.~Wang.
\newblock Pose-aware adversarial domain adaptation for personalized facial expression recognition.
\newblock {\em arXiv preprint arXiv:2007.05932}, 2020.

\bibitem{liu2023evaef}
X.~Liu, L.~Sun, W.~Jiang, F.~Zhang, Y.~Deng, Z.~Huang, L.~Meng, Y.~Liu, and C.~Liu.
\newblock Evaef: Ensemble valence-arousal estimation framework in the wild.
\newblock In {\em Proceedings of the IEEE/CVF Conference on Computer Vision and Pattern Recognition}, pages 5862--5870, 2023.

\bibitem{lopez2018multi}
D.~Lopez-Martinez, K.~Peng, S.~C. Steele, A.~J. Lee, D.~Borsook, and R.~Picard.
\newblock Multi-task multiple kernel machines for personalized pain recognition from functional near-infrared spectroscopy brain signals.
\newblock In {\em 2018 24th International Conference on Pattern Recognition (ICPR)}, pages 2320--2325. IEEE, 2018.

\bibitem{malkauthekar2013analysis}
M.~Malkauthekar.
\newblock Analysis of euclidean distance and manhattan distance measure in face recognition.
\newblock In {\em Third International Conference on Computational Intelligence and Information Technology (CIIT 2013)}, pages 503--507. IET, 2013.

\bibitem{niinuma2022facial}
K.~Niinuma, I.~{\"O}nal~Ertu{\u{g}}rul, J.~F. Cohn, L.~A. Jeni, et~al.
\newblock Facial expression manipulation for personalized facial action estimation.
\newblock {\em Frontiers in Signal Processing}, 2:1--16, 2022.

\bibitem{odaka2023block}
Y.~Odaka and K.~Kaneiwa.
\newblock Block-segmentation vectors for arousal prediction using semi-supervised learning.
\newblock {\em Applied Soft Computing}, 142:110327, 2023.

\bibitem{poria2016convolutional}
S.~Poria, I.~Chaturvedi, E.~Cambria, and A.~Hussain.
\newblock Convolutional mkl based multimodal emotion recognition and sentiment analysis.
\newblock In {\em 2016 IEEE 16th international conference on data mining (ICDM)}, pages 439--448. IEEE, 2016.

\bibitem{povolny2016multimodal}
F.~Povolny, P.~Matejka, M.~Hradis, A.~Popkov{\'a}, L.~Otrusina, P.~Smrz, I.~Wood, C.~Robin, and L.~Lamel.
\newblock Multimodal emotion recognition for avec 2016 challenge.
\newblock In {\em Proceedings of the 6th International Workshop on Audio/Visual Emotion Challenge}, pages 75--82, 2016.

\bibitem{praveen2023audio}
R.~G. Praveen, P.~Cardinal, and E.~Granger.
\newblock Audio-visual fusion for emotion recognition in the valence-arousal space using joint cross-attention.
\newblock {\em IEEE Transactions on Biometrics, Behavior, and Identity Science}, 2023.

\bibitem{ps2017emotion}
S.~PS and G.~Mahalakshmi.
\newblock Emotion models: a review.
\newblock {\em International Journal of Control Theory and Applications}, 10(8):651--657, 2017.

\bibitem{pujol2019emotion}
F.~A. Pujol, H.~Mora, and A.~Mart{\'\i}nez.
\newblock Emotion recognition to improve e-healthcare systems in smart cities.
\newblock In {\em Research \& Innovation Forum 2019: Technology, Innovation, Education, and their Social Impact 1}, pages 245--254. Springer, 2019.

\bibitem{ren2022predicting}
B.~Ren, E.~G. Balkind, B.~Pastro, E.~S. Israel, D.~A. Pizzagalli, H.~Rahimi-Eichi, J.~T. Baker, and C.~A. Webb.
\newblock Predicting states of elevated negative affect in adolescents from smartphone sensors: A novel personalized machine learning approach.
\newblock {\em Psychological Medicine}, pages 1--9, 2022.

\bibitem{Rescigno202035811}
M.~Rescigno, M.~Spezialetti, and S.~Rossi.
\newblock Personalized models for facial emotion recognition through transfer learning.
\newblock {\em Multimedia Tools and Applications}, 79(47-48):35811 – 35828, 2020.
\newblock Cited by: 12; All Open Access, Hybrid Gold Open Access.

\bibitem{rescigno2020personalized}
M.~Rescigno, M.~Spezialetti, and S.~Rossi.
\newblock Personalized models for facial emotion recognition through transfer learning.
\newblock {\em Multimedia Tools and Applications}, 79:35811--35828, 2020.

\bibitem{rudovic2018personalized}
O.~Rudovic, J.~Lee, M.~Dai, B.~Schuller, and R.~W. Picard.
\newblock Personalized machine learning for robot perception of affect and engagement in autism therapy.
\newblock {\em Science Robotics}, 3(19), 2018.

\bibitem{saeed2017personalized}
A.~Saeed and S.~Trajanovski.
\newblock Personalized driver stress detection with multi-task neural networks using physiological signals.
\newblock {\em arXiv preprint arXiv:1711.06116}, 2017.

\bibitem{salam2023automatic}
H.~Salam, O.~Celiktutan, H.~Gunes, and M.~Chetouani.
\newblock Automatic context-aware inference of engagement in hmi: A survey.
\newblock {\em IEEE Transactions on Affective Computing}, 2023.

\bibitem{salam2016fully}
H.~Salam, O.~Celiktutan, I.~Hupont, H.~Gunes, and M.~Chetouani.
\newblock Fully automatic analysis of engagement and its relationship to personality in human-robot interactions.
\newblock {\em IEEE Access}, 5:705--721, 2016.

\bibitem{team2iccv}
H.~Salam, O.~Celiktutan, V.~Manoranjan, I.~Ismail, and H.~Mukherjee.
\newblock Learning personalised models for automatic self-reported personality recognition.
\newblock In {\em ICCV 2021 Understanding Social Behavior in Dyadic and Small Group Interactions Challenge Fact sheet: Automatic self-reported personality recognition Track}, 2021.

\bibitem{salam2015engagement}
H.~Salam and M.~Chetouani.
\newblock Engagement detection based on mutli-party cues for human robot interaction.
\newblock In {\em 2015 International Conference on Affective Computing and Intelligent Interaction (ACII)}, pages 341--347. IEEE, 2015.

\bibitem{san2017euclidean}
E.~San~Segundo, A.~Tsanas, and P.~G{\'o}mez-Vilda.
\newblock Euclidean distances as measures of speaker similarity including identical twin pairs: a forensic investigation using source and filter voice characteristics.
\newblock {\em Forensic Science International}, 270:25--38, 2017.

\bibitem{savchenko2023emotieffnets}
A.~V. Savchenko.
\newblock Emotieffnets for facial processing in video-based valence-arousal prediction, expression classification and action unit detection.
\newblock In {\em Proceedings of the IEEE/CVF Conference on Computer Vision and Pattern Recognition}, pages 5715--5723, 2023.

\bibitem{schneider2021personalization}
J.~Schneider and M.~Vlachos.
\newblock Personalization of deep learning.
\newblock In {\em Data Science--Analytics and Applications: Proceedings of the 3rd International Data Science Conference--iDSC2020}, pages 89--96. Springer, 2021.

\bibitem{schroff2015facenet}
F.~Schroff, D.~Kalenichenko, and J.~Philbin.
\newblock Facenet: A unified embedding for face recognition and clustering.
\newblock In {\em Proceedings of the IEEE conference on computer vision and pattern recognition}, pages 815--823, 2015.

\bibitem{senechal2012facial}
T.~Senechal, V.~Rapp, H.~Salam, R.~Seguier, K.~Bailly, and L.~Prevost.
\newblock Facial action recognition combining heterogeneous features via multikernel learning.
\newblock {\em IEEE Transactions on Systems, Man, and Cybernetics, Part B (Cybernetics)}, 42(4):993--1005, 2012.

\bibitem{shah2021personalized}
R.~V. Shah, G.~Grennan, M.~Zafar-Khan, F.~Alim, S.~Dey, D.~Ramanathan, and J.~Mishra.
\newblock Personalized machine learning of depressed mood using wearables.
\newblock {\em Translational psychiatry}, 11(1):1--18, 2021.

\bibitem{Shahabinejad2021toward}
M.~Shahabinejad, Y.~Wang, Y.~Yu, J.~Tang, and J.~Li.
\newblock Toward personalized emotion recognition: A face recognition based attention method for facial emotion recognition.
\newblock In {\em Proceedings of IEEE International Conference on Face \& Gesture}, 2021.

\bibitem{shao2021personality}
Z.~Shao, S.~Song, S.~Jaiswal, L.~Shen, M.~Valstar, and H.~Gunes.
\newblock Personality recognition by modelling person-specific cognitive processes using graph representation.
\newblock In {\em proceedings of the 29th ACM international conference on multimedia}, pages 357--366, 2021.

\bibitem{shaqra2019recognizing}
F.~A. Shaqra, R.~Duwairi, and M.~Al-Ayyoub.
\newblock Recognizing emotion from speech based on age and gender using hierarchical models.
\newblock {\em Procedia Computer Science}, 151:37--44, 2019.

\bibitem{siirtola2023predicting}
P.~Siirtola, S.~Tamminen, G.~Chandra, A.~Ihalapathirana, and J.~R{\"o}ning.
\newblock Predicting emotion with biosignals: A comparison of classification and regression models for estimating valence and arousal level using wearable sensors.
\newblock {\em Sensors}, 23(3):1598, 2023.

\bibitem{singh2019speaker}
M.~K. Singh, N.~Singh, and A.~Singh.
\newblock Speaker's voice characteristics and similarity measurement using euclidean distances.
\newblock In {\em 2019 International Conference on Signal Processing and Communication (ICSC)}, pages 317--322. IEEE, 2019.

\bibitem{soladie2012multimodal}
C.~Soladi{\'e}, H.~Salam, C.~Pelachaud, N.~Stoiber, and R.~S{\'e}guier.
\newblock A multimodal fuzzy inference system using a continuous facial expression representation for emotion detection.
\newblock In {\em Proceedings of the 14th ACM international conference on Multimodal interaction}, pages 493--500, 2012.

\bibitem{soladie2013continuous}
C.~Soladi{\'e}, H.~Salam, N.~Stoiber, and R.~S{\'e}guier.
\newblock Continuous facial expression representation for multimodal emotion detection.
\newblock {\em International Journal of Advanced Computer Science (IJACSci)}, 3(5), 2013.

\bibitem{song2021hidden}
B.~C. Song and D.~H. Kim.
\newblock Hidden emotion detection using multi-modal signals.
\newblock In {\em Extended Abstracts of the 2021 CHI Conference on Human Factors in Computing Systems}, pages 1--7, 2021.

\bibitem{song2016perceived}
Y.~Song, S.~Dixon, M.~T. Pearce, and A.~R. Halpern.
\newblock Perceived and induced emotion responses to popular music: Categorical and dimensional models.
\newblock {\em Music Perception: An Interdisciplinary Journal}, 33(4):472--492, 2016.

\bibitem{stappen2021muse}
L.~Stappen, A.~Baird, L.~Christ, L.~Schumann, B.~Sertolli, E.-M. Messner, E.~Cambria, G.~Zhao, and B.~W. Schuller.
\newblock The muse 2021 multimodal sentiment analysis challenge: sentiment, emotion, physiological-emotion, and stress.
\newblock In {\em Proceedings of the 2nd on Multimodal Sentiment Analysis Challenge}, pages 5--14. 2021.

\bibitem{stappen2021muse2}
L.~Stappen, E.-M. Me{\ss}ner, E.~Cambria, G.~Zhao, and B.~W. Schuller.
\newblock Muse 2021 challenge: Multimodal emotion, sentiment, physiological-emotion, and stress detection.
\newblock In {\em Proceedings of the 29th ACM International Conference on Multimedia}, pages 5706--5707, 2021.

\bibitem{taylor2017personalized}
S.~Taylor, N.~Jaques, E.~Nosakhare, A.~Sano, and R.~Picard.
\newblock Personalized multitask learning for predicting tomorrow's mood, stress, and health.
\newblock {\em IEEE Transactions on Affective Computing}, 11(2):200--213, 2017.

\bibitem{vaswani2017attention}
A.~Vaswani, N.~Shazeer, N.~Parmar, J.~Uszkoreit, L.~Jones, A.~N. Gomez, {\L}.~Kaiser, and I.~Polosukhin.
\newblock Attention is all you need.
\newblock {\em Advances in neural information processing systems}, 30, 2017.

\bibitem{vijaymeena2016survey}
M.~Vijaymeena and K.~Kavitha.
\newblock A survey on similarity measures in text mining.
\newblock {\em Machine Learning and Applications: An International Journal}, 3(2):19--28, 2016.

\bibitem{vogt-andre-2006-improving}
T.~Vogt and E.~Andr{\'e}.
\newblock Improving automatic emotion recognition from speech via gender differentiaion.
\newblock In {\em Proceedings of the Fifth International Conference on Language Resources and Evaluation ({LREC}{'}06)}, Genoa, Italy, May 2006. European Language Resources Association (ELRA).

\bibitem{wagner2023dawn}
J.~Wagner, A.~Triantafyllopoulos, H.~Wierstorf, M.~Schmitt, F.~Burkhardt, F.~Eyben, and B.~W. Schuller.
\newblock Dawn of the transformer era in speech emotion recognition: closing the valence gap.
\newblock {\em IEEE Transactions on Pattern Analysis and Machine Intelligence}, 2023.

\bibitem{wang2018personalized}
C.~Wang and S.~Wang.
\newblock Personalized multiple facial action unit recognition through generative adversarial recognition network.
\newblock In {\em Proceedings of the 26th ACM international conference on Multimedia}, pages 302--310, 2018.

\bibitem{woodward2021towards}
K.~Woodward, E.~Kanjo, D.~J. Brown, and T.~McGinnity.
\newblock Towards personalised mental wellbeing recognition on-device using transfer learning “in the wild”.
\newblock In {\em 2021 IEEE International Smart Cities Conference (ISC2)}, pages 1--7. IEEE, 2021.

\bibitem{yang2018identity}
H.~Yang, Z.~Zhang, and L.~Yin.
\newblock Identity-adaptive facial expression recognition through expression regeneration using conditional generative adversarial networks.
\newblock In {\em 2018 13th IEEE International Conference on Automatic Face \& Gesture Recognition (FG 2018)}, pages 294--301. IEEE, 2018.

\bibitem{yusuf2017individuality}
R.~Yusuf, D.~G. Sharma, I.~Tanev, and K.~Shimohara.
\newblock Individuality and user-specific approach in adaptive emotion recognition model.
\newblock In {\em 2017 International Conference on Biometrics and Kansei Engineering (ICBAKE)}, pages 1--6. IEEE, 2017.

\bibitem{zhao2018transferring}
H.~Zhao, N.~Ye, and R.~Wang.
\newblock Transferring age and gender attributes for dimensional emotion prediction from big speech data using hierarchical deep learning.
\newblock In {\em 2018 IEEE 4th International Conference on Big Data Security on Cloud (BigDataSecurity), IEEE International Conference on High Performance and Smart Computing,(HPSC) and IEEE International Conference on Intelligent Data and Security (IDS)}, pages 20--24. IEEE, 2018.

\bibitem{zhao2019personalized}
S.~Zhao, A.~Gholaminejad, G.~Ding, Y.~Gao, J.~Han, and K.~Keutzer.
\newblock Personalized emotion recognition by personality-aware high-order learning of physiological signals.
\newblock {\em ACM Transactions on Multimedia Computing, Communications, and Applications (TOMM)}, 15(1s):1--18, 2019.

\bibitem{zhi2020comprehensive}
R.~Zhi, M.~Liu, and D.~Zhang.
\newblock A comprehensive survey on automatic facial action unit analysis.
\newblock {\em The Visual Computer}, 36:1067--1093, 2020.

\end{thebibliography}
}

\end{document}